\documentclass[conference]{IEEEtran}
\IEEEoverridecommandlockouts
\usepackage[ruled,vlined,linesnumbered]{algorithm2e}
\usepackage{cite}
\usepackage{amsmath,amssymb,amsfonts,algorithmic,textcomp, multirow}
\usepackage{graphicx,subfig,float,xcolor,url,hyperref,etoolbox}
\usepackage[utf8]{inputenc}
\usepackage[T1]{fontenc}
\usepackage[nolist]{acronym}
\hypersetup{
    colorlinks,
    linkcolor={blue!50!black},
    citecolor={blue!50!black},
    urlcolor={black!80!black}
}

\def\BibTeX{{\rm B\kern-.05em{\sc i\kern-.025em b}\kern-.08em
    T\kern-.1667em\lower.7ex\hbox{E}\kern-.125emX}}

\makeatletter
  \patchcmd{\@maketitle}
   {\addvspace{0.5\baselineskip}\egroup}
   {\addvspace{-0.1\baselineskip}\egroup}
   {}
   {}
\makeatother

\begin{document}
\title{Learning Grasp Affordance Reasoning through Semantic Relations}
\author{Paola Ard{\'o}n$^{*}$, {\`E}ric Pairet$^{*}$, Ronald P. A. Petrick$^{\dagger}$, Subramanian Ramamoorthy$^{*}$, and Katrin S. Lohan$^{\dagger}$
\thanks{Edinburgh Centre for Robotics. University of Edinburgh and Heriot-Watt University. Edinburgh, UK. $^{\dagger}$\{r.petrick;k.lohan\}@hw.ac.uk}
\thanks{$^{*}$\{paola.ardon;eric.pairet;s.ramamoorthy\}@ed.ac.uk;}
}

\begin{acronym}[ransac]
  \acro{LbD}{learning by demonstration}
  \acro{RL}{reinforcement learning}
  \acro{SVM}{Support Vector Machine}
  \acro{DOF}{degrees-of-freedom}
  \acro{CAD}{computer-aided design}
  \acro{ROI}{regions of interest}
  \acro{MRF}{Markov Random Fields}
  \acro{ECV}{early cognitive vision}
  \acro{IADL}{instrumental activities of daily living}
  \acro{CDR}{cognitive developmental robotics}
  \acro{2-D}{two-dimensional}
  \acro{3-D}{three-dimensional}
  \acro{RANSAC}{Random sample consensus}
  \acro{RGB-D}{red-green-blue depth}
  \acro{IFR}{International Federation of Robotics}
  \acro{CNN}{Convolutional Neural Network}
  \acro{KB}{knowledge base}
  \acro{MSE}{mean square error}
  \acro{MLN}{Markov Logic Network}
  \acro{XAI}{Explainable Artificial Intelligence}
  \acro{MC-SAT}{Model-Constructing Satisfiability Calculus}
  \acro{WCSP}{weighted constraint satisfaction problem}
  \acro{MAP}{Maximum--Likelihood}
  \acro{O-CNN}{Octree-based Convolutional Neural Networks}
  \acro{OACs}{object-action complexes}
  \acro{CAD}{computer-aided-design}
  \acro{ROC}{Receiver Operating Characteristics}
  \acro{AUC}{area under the curve}
  \acro{MCMC}{Markov chain Monte Carlo}
  \acro{FOL}{first-order logic}
  \end{acronym}

\maketitle

\begin{abstract}
Reasoning about object affordances allows an autonomous agent to perform generalised manipulation tasks among object instances.
While current approaches to grasp affordance estimation are effective, they are limited to a single hypothesis.
We present an approach for detection and extraction of multiple grasp affordances on an object via visual input.
We define semantics as a combination of multiple attributes, which yields benefits in terms of generalisation for grasp affordance prediction.
We use Markov Logic Networks to build a knowledge base graph representation to obtain a probability distribution of grasp affordances for an object.
To harvest the knowledge base, we collect and make available a novel dataset that relates different semantic attributes.
We achieve reliable mappings of the predicted grasp affordances on the object by learning prototypical grasping patches from several examples.
We show our method's generalisation capabilities on grasp affordance prediction for novel instances and compare with similar methods in the literature. Moreover, using a robotic platform, on simulated and real scenarios, we evaluate the success of the grasping task when conditioned on the grasp affordance prediction.

\end{abstract}

\section{Introduction}\label{sec:intro}

Modern robotic platforms are capable of performing a rich set of human-scale manipulation tasks. Affordance is one of the key concepts that enables an autonomous agent to interact with a variety of objects successfully. Affordance refers to the possibility of performing different actions with an object \cite{Gibson77-affordances}.
By associating context and previous experiences, humans are very effective at creating grasp affordance relations to facilitate an intended action. 
For example, grasping a pair of scissors from the tip affords handing over the tool, but not a cutting task.
In the context of robotics, grasp affordances have attained new relevance as agents should be able to manipulate novel objects for tasks with distinct contextualisations. 

The current literature offers solutions that are successful in real-world scenarios, but typically assign a single universal grasp affordance for a given object no matter the context of the scene \cite{Montesano2009LearningGA,bonaiuto2015learning,Hart2015TheAT,AffordanceNet18}. In reality, a single object affords different actions, and the successful accomplishment of the task is dependant on identifying the correct grasping region of the object. Nonetheless, in robotics, there is a relational gap between the interaction of different object categories associated with changing scenarios and pose-grasps. 
The missed connection between objects and grasp relationships has resulted in (i)~a tendency only to consider a single grasp affordance per object, and (ii)~a lack of datasets that take into account the relational aspects of grasp affordance.
\begin{figure}[t!]
  \centering
  \includegraphics[width= 8.7cm]{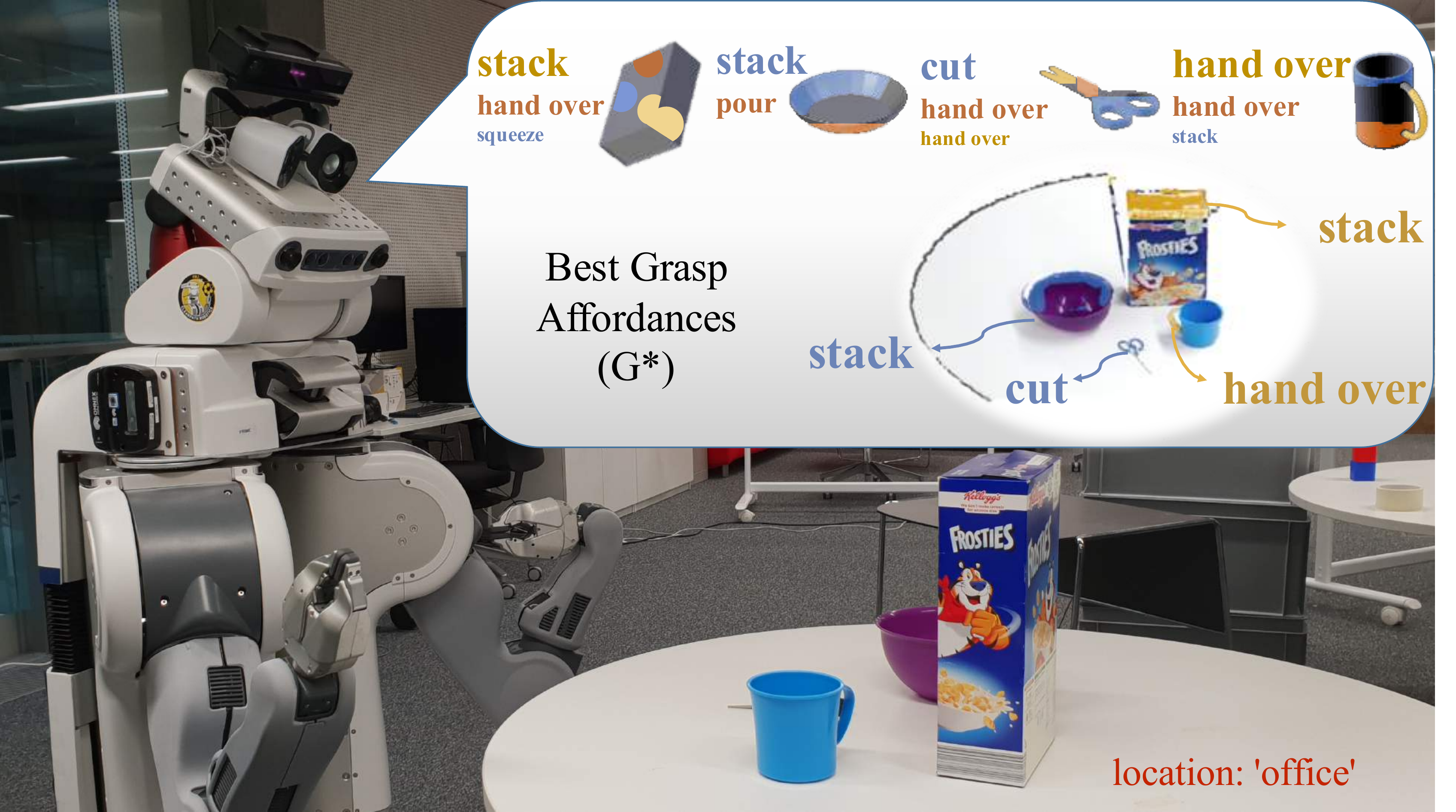}
  \vspace{-0.15cm}
  \caption{PR2 reasoning about grasp affordances of objects on a tabletop office scenario. The affordances are colour coded with the corresponding grasping region on the objects. On the top three affordance labels, the larger the size, the more suitable that affordance is in the perceived context.\label{fig:intro}}
  \vspace{-0.2cm}
\end{figure}
On the grounds of the limitations mentioned above, the contribution of our approach is threefold.
First, we present an approach for multi-target prediction of grasp affordances on an object using \ac{MLN} theory to build a \ac{KB} graph representation. Our method is able to reason about the most probable grasp affordance, among a set, by inferring the context semantics relation using Gibbs sampling. 
Second, to test the prediction on the grasping task, we map the obtained grasp affordances to the \ac{3-D} data of the object. The system learns the object shape context and related prototypical grasping patches to create hypotheses of grasp locations. The most probable grasp affordance is then chosen to generate a reaching and grasping configuration plan.
Finally, we collect and make available a new dataset for visual grasp affordance prediction\footnote{Data and code: \url{https://paolaardon.github.io/grasp_affordance_reasoning/}\label{ft:data}} that promotes more robust and heterogeneous robotic grasping methods. The dataset contains different attributes from $30$ different object classes. Each instance is related not only to the semantic descriptions, but also to the physical features describing visual attributes, locations, and different grasping regions for a variety of actions.

In addition, we also compare the generalisation of the grasp affordance predictions on novel objects against 
current state-of-the-art techniques. The reliability of the obtained grasp affordance regions is evaluated using similarity metrics. We compare these calculated hypotheses with the ground truth labels obtaining high correlation values.
We analyse how feasible our approach is for a general tabletop scenario, as shown in Fig.~\ref{fig:intro}, with known and novel objects in simulated and real indoor scenes.
\begin{figure*}[!hbt]
  \centering
  \includegraphics[width=16cm]{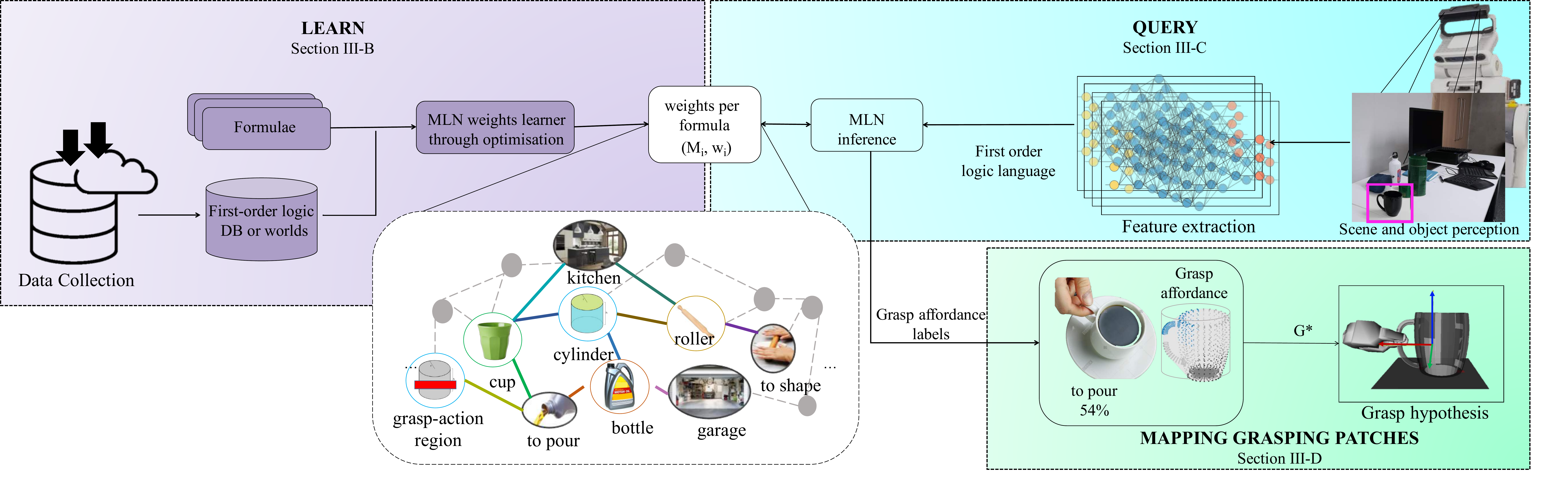}
  \caption{Proposed framework for reasoning about object grasp affordances, composed of the learning, querying and mapping tasks. The learnt model (white box) encodes the relation between nodes with connecting coloured edges.}\label{fig:framework}
\end{figure*}

\section{Related Work}\label{sec:related_work}

Learning visual grasp affordances for objects has been an active area of research for robotic manipulation tasks. 
Ideally, an autonomous agent should be able to distinguish between objects and their utility. In real-world scenarios, a single object affords different actions, and the successful completion of the task is dependent on the correct choice of the grasping region. The literature on visually detecting robotic graspings is vast. The state-of-the-art in this area, \cite{Lenz2015,bohg2010learning}, offers robust methodologies for identifying candidate grasps, either with architectures based on deep learning to detect grasp areas on an object \cite{Lenz2015}, or using supervised learning to obtain grasping points based on objects shape \cite{bohg2010learning}. While these techniques offer robust grasp candidates, they do not differentiate among actions at those detected grasp points.

\subsubsection*{\textbf{Grasp affordances}}
Work on grasp affordances tends to focus on robust interactions between objects and the autonomous agent. However, it is typically limited to a single affordance per object. Moreover, affordance labels tend to be assigned arbitrarily instead of through data-driven techniques gathering human judgement to portray socially acceptable interactions regarding the grasps.
Some works, such as \cite{kruger2011object}, focus on relating abstractions of sensory-motor processes with object structures (e.g., \ac{OACs}) to extract the best reaching and grasping candidate given an object affordance. Others use purely visual input to learn affordances using deep learning \cite{Nguyen17,AffordanceNet18} or supervised learning techniques to relate objects and actions \cite{song2010learning,Hart2015TheAT,bonaiuto2015learning,Montesano2009LearningGA,moldovan2012learning,detry2009learning}.
In contrast, our approach reasons about a set of affordance possibilities for an object using data-driven techniques to discern the best grasping approach to succeed at a given action task.

\subsubsection*{\textbf{Datasets}}
At present, no dataset offers an end-to-end relation between objects and grasp affordances.
Some datasets relate objects with actions for affordance detection \cite{Nguyen17,AffordanceNet18}. Others offer a mapping of robust grasps labels for different objects without considering the actions \cite{jiang2011efficient}. Not having a dataset that brings together both concepts represents a problem for any methodology that aims to achieve a robust social human-robot interaction architecture for manipulation tasks. In contrast, we harvest a dataset that includes object locations, grasp labels and semantics as relational attributes, encouraging more robust and heterogeneous robotic grasping methods.

\subsubsection*{\textbf{Knowledge bases (KB)}}

A \acl{KB} refers to a repository of entities and rules that can be used for storing and querying objects' affordance information. 
\acp{MLN}~\cite{richardson2006markov} represent the current state-of-the-art method when it comes to reasoning about objects using knowledge bases.  
An \ac{MLN} is a combination of \ac{MRF} with a \ac{FOL} language. \cite{zhu2014reasoning} uses \ac{MLN} to learn the optimal weights that relate different object descriptions extracted with a ranking \ac{SVM} function. On the other hand, \cite{farhadi2009describing} trains a battery of L1-linear regularised logistic classifiers and learn the attributes relation according to the classification score.
The performance of the \acp{KB} using \ac{MLN} has been shown to outperform alternatives \cite{fadiga2000visuomotor,ardon2018Towards} given the Markovian ability to relate attributes. Using \ac{MLN} in \acp{KB} is advantageous as it can incorporate the uncertainty of probabilistic graphical models. This performance depends on the quality and correlation of the data used for training.

In contrast to current approaches in the field, we collect a detailed dataset that promotes robust grasp techniques. We use this dataset to reason about an object's multiple reliable grasps, corresponding to actions resulting from the design of a \ac{KB} based on \ac{MLN}.

\section{Proposed Method \label{sec:method}}

Our primary task is to reason about feasible grasps in an object that are closely related to the success of an affordance task. The grasp affordance relationship is built using semantics as a collection of attributes (as explained in Section~\ref{sc:gathering}). Fig.~\ref{fig:framework} shows a summary of our proposed methodology as follows: (i)~we learn the semantics relation between attributes, locations and grasp affordances through a unique building of grounding and combination of rules using \ac{MLN}, (ii)~we query an approximation of the probability distribution associated with grasp affordances using Gibbs sampling \cite{liu1989limited}, and (iii)~among all the possibilities, we take the one that satisfies a given affordance with the highest probability. This selected grasp affordance region is then located on the \ac{3-D} object data to calculate a grasping configuration.

\subsection{Knowledge Base Terminology\label{sc:KB}}

A \acl{KB} can be represented as a graph, similar to the white block in Fig.~\ref{fig:framework}. The nodes denote the entities and the edges the general rules that characterise their relationship. For example, a cup is a node or entity connected to other nodes depicting its visual attributes, its affordance (such as \textit{pour}) and the corresponding grasping region. These entities are connected with edges of different colours representing the different weights. The higher the weight, the more likely that relation is to be true.
We build the \ac{KB} by learning these relations, i.e. the weights of the general rules. We employ an \ac{MLN} \cite{richardson2006markov} for knowledge representation. To construct a \ac{KB} with \ac{MLN} there is a pre-learning process: the first step is to collect evidence (as detailed in Section~\ref{sc:gathering}), in the form of a set of facts and assertions about the entities. The different sets of assertions create possible \textit{worlds}. For example, scissors have metal blades and handles. These two assertions create a world where objects having these two characteristics are likely to be a pair of scissors. The second step is to create a general set of rules. Each of these rules is a \textit{formula} $M_i$ associated with a \textit{weight} $w_i$, thus creating correlated pairs \mbox{($M_i$, $w_i$)}. 
The formulae are built by creating a relation between the entities. In an \ac{MLN}, the entities are \textit{terms} and the relation between terms are \textit{predicates}. Table~\ref{tb:examples_terminology} shows examples of possible \textit{predicates} and \textit{formulae} in our \ac{KB}. For example, the \textit{predicate} ``hasShape" is a relation between the terms ``object" and ``shape".
All the terms, except ``object", are considered \textit{grounded terms} since we know their domain representation.

\subsection{Learning Grasp Affordance Relations} \label{sec:method_semantic_relations}

The possible \textit{worlds} $x$ that we collect (Section~\ref{sc:gathering}) are used for learning the formulae' weights and are translated into \ac{FOL} predicates to form formula sets $M$. Table~\ref{tb:examples_terminology} shows some examples of allowable combinations of predicates used in our \ac{KB}. The location, category, grasping region and visual attributes (i.e., texture, material, shape) are treated as constants and the object as a variable.
Given the different sets of constants inside each term, different networks are produced. These networks are of widely varying sizes, but all grounded terms of each formula $M_i$ have the same weight $w_i$.
The weights are then learned generatively using the available possible \textit{worlds} $x$ by calculating their joint distribution as:

\begin{equation}
    P(X=x) = 
    \frac{1}{Z} \exp\Big(\sum_{i=1}^n w_i f_i(x_{\{i\}})\Big),
\end{equation}
where $Z$ is the normalisation constant over the potential functions $\phi_i$ of connected nodes given by $\sum_{x \in \mathcal{X}} \prod_{i} \phi_{i}(x_{\{i\}})$, $n$ is the number of formulae in $M$, $x_{\{i\}}$ is the state of the grounded terms (i.e., the state of existing terms that appear in that world) in $M_i$, and the feature function $f_i(x_{\{i\}})=1$ if $M_i(x_{\{i\}})$ is true or $0$ otherwise. The weights $w_i$ indicate the likelihood of the formula being true.
Using Broyden's method \cite{liu1989limited} we learn the optimal weights $w^*$ from maximising the pseudo-log-likelihood \mbox{$log P^*_w(X=x)$} of the obtained probability distribution of the available worlds. Table~\ref{tb:examples_terminology} shows some examples of the learned relations ($M_i, w_i$) in the \ac{KB}. For example, a container is ten times more likely to afford a pouring task than an object categorised as electronic. 

\begin{table}[t!]
    \centering
    \begin{tabular}{p{4.5cm}| p{3.5cm} }
    \hline
    \multicolumn{2}{c}{\textbf{LEARNING}} \\ \cline{1-2} 
    \textbf{Predicates} &  \textbf{Formula-weights ($M_i, w_i$)}  \\ \hline 
    \textbf{hS:} hasShape(obj, shape) & \\
    \textbf{hT:} hasTexture(obj, texture) & $(\text{hS}\wedge \text{hT}\wedge \text{hM}\Rightarrow \text{hA}, w_1)$ \\
    
    \textbf{hM:} hasMaterial(obj, material) & ($\text{hS}\wedge \text{hT}\wedge \text{hM}\wedge \text{cF}\Rightarrow \text{hA}, w_2$) \\
    
    \textbf{cF:} canBeFound(obj, location) &  ...\\
    
    \textbf{hA:} hasAffordance(obj, affordance) & ($\text{hC}\Rightarrow \text{hA},w_{n-1}$)\\
    \textbf{hC:} hasCategory(obj, category) & ($ \text{hA}\Rightarrow \text{gR},w_{n}$ ) \\
    \textbf{gR:} graspRegion(obj, region) &   \\ \hline

   \multicolumn{2}{c}{\textbf{EXAMPLES}} \\ \cline{1-2}
   \multicolumn{2}{c}{($\text{hasCategory(obj, container)}\Rightarrow \text{hasAffordance(obj, pour)}, log(0.67)$)} \\ 
   \multicolumn{2}{c}{($\text{hasCategory(obj, electronics)}\Rightarrow \text{hasAffordance(obj, pour)}, log(0.07)$)} \\ 
   
   \hline
    \end{tabular}
    \caption{Knowledge base schema for the learning task. Our formulae are defined as relations between predicates. The examples give an idea of the learned relations with weights. \label{tb:examples_terminology}}
    \vspace{-0.25cm}
 \end{table}


\begin{table}[b!]
     \centering
     \begin{tabular}{p{4.5cm}| p{3.5cm} }
     \hline
   
   \multicolumn{2}{c}{\textbf{QUERY,} given the attributes of a cup:} \\ \cline{1-2}
   \hline
   \multicolumn{2}{l}{$\text{hasAffordance(obj, x)}\wedge \text{graspRegion(obj, x)}$} \\ 
   
   \multicolumn{2}{l}{($\text{hasAffordance(obj, stack)}\wedge \text{graspRegion(obj, 1)}, 49\%)$} \\ 
   \multicolumn{2}{l}{($\text{hasAffordance(obj, hand over)}\wedge \text{graspRegion(obj, 3)}, 17\%)$} \\ 
   \multicolumn{2}{l}{($\text{hasAffordance(obj, pour)}\wedge \text{graspRegion(obj, 2)}, 22\%)$} \\ 
    \hline
    
     \end{tabular}
     \caption{Example of a query and the top answers given an object's attributes presented as assertions that build a world. \label{tb:examples_query}}
 \end{table}

\subsection{Reasoning about Grasp Affordances} \label{sec:method_labels}
To reason about an object's grasp affordances, we use the \ac{2-D} image and pass it through a deep learning architecture built with pre-trained \acp{CNN}~\cite{he2016deep} to extract the objects' attributes (i.e., shape, texture, material, category, location) as labels. These labels are translated into \ac{FOL} to create possible worlds. These worlds then serve to query the most feasible grasping region for an affordance task. Table~\ref{tb:examples_query} shows an example of probable relations between affordance and grasping region labels (i.e., 1, 2 or 3) given a grasp affordance query. In the \ac{KB} the more formulae a world adheres to, the more probable it is.
In order to query grasp affordances from the learned weights model, we use Gibbs sampling \cite{kim1999state}.
We employ Gibbs sampling to generate posterior samples by sweeping through each grounded term while keeping the calculations tractable. We compute the expectation of a posterior distribution as:
\begin{equation}
E[h(s)]_\mathcal{P} \approx \frac{1}{n}\sum_{i=1}^n h(s^{(i)}),
\end{equation}
where $n$ is the number of grounded simulated samples from that distribution, $\mathcal{P}$ is the posterior distribution of the world of interest, $h(s)$ is the desired expectation and $h(s^{(i)})$ is the $i^{th}$ simulated sample from $\mathcal{P}$. 
This inference method gives us the maximum probability for different grasping regions for every query $q \in \boldsymbol{Q}$. The corresponding labels (see example in Table II) are used to map the resulting grasping affordances on the object \ac{3-D} data. Among these grasp affordances, the most likely one is used as the optimal grasping patch label $G^*=\arg \max_n E[h(s)]$ on which we calculate a grasping configuration to be sent to the robot.

\subsection{Mapping Grasp Affordance Patches} \label{sec:method_regions}
\begin{figure}[t]
   \centering
   \includegraphics[width=6.3cm]{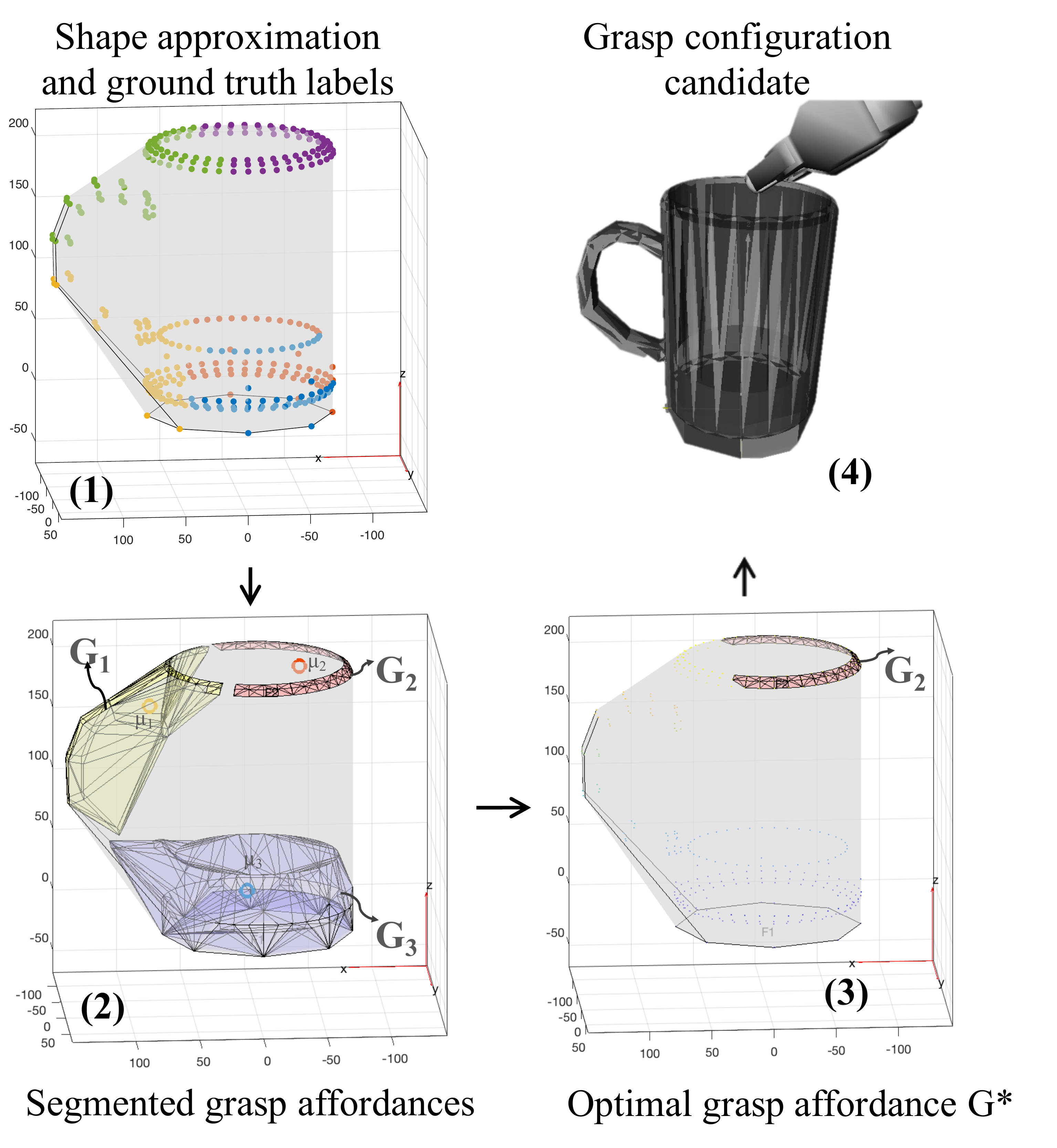}
   \caption{Mapping grasp affordance patches on \ac{3-D} object data.\label{fig:grasp}}
 \end{figure}

Our goal is to produce robust manipulation by reasoning about the best possible grasp for a given affordance. Towards this objective, we map the previously obtained grasp affordance probabilities to the object's \ac{3-D} data. We use a combination of the object shape context and hierarchical segmentation of the point cloud.
To ensure the grasping regions on the object are reliable, we adopt as a starting point the pre-defined grasping regions from \cite{jiang2011efficient} (\ac{2-D} with corresponding \ac{3-D} mapping) and use them as ground truth labels. Nonetheless, after parsing the data collected in Section~\ref{sc:gathering}, not all the ground truth labels (colour coded labels from \cite{jiang2011efficient} in Fig.~\ref{fig:grasp}, step \textbf{(1)}) afford an action or different labels could afford the same action. Hence, we redefine sets of ground truth labels into our grasp affordance patches. 
Using the ground truth data in \cite{jiang2011efficient}, the points are clustered using the k-means algorithm, extracting $n$ grasping regions (as probabilities obtained from the \ac{KB} querying step). 
The cluster centres $\mu_G$ serve as the seeds for the representative initial patches $G$. 
A set of faces forms these patches. The faces are grouped using hierarchical mesh decomposition as proposed in \cite{katz2003hierarchical}. The mesh decomposition produces $G_n$ grasping patches as shown in step \textbf{(2)} in Fig.~\ref{fig:grasp}. We consider the features belonging to the $G_n$ grasping patches as inputs and classify them in $n$ grasp affordance candidate labels based on the object's context shape using an \ac{SVM} classifier as done in \cite{Lenz2015,bohg2010learning}.
The optimal grasping patch label $G^*$ is represented by a set $\boldsymbol{b}$ of \ac{3-D} points which have a dominant plane $\widehat{\Pi}$ with centroid $\boldsymbol{\nu_C}$ and orientation $\boldsymbol{\gamma}$ that serves as the position and orientation for the inverse kinematics calculation of the grasping approach (Fig.~\ref{fig:grasp}, step \textbf{(3)} and \textbf{(4)}).

\subsection{End-to-end Execution}

Algorithm \ref{alg:framework_kb} presents an outline of the framework's end-to-end execution, which aims to provide a robotic platform with a feasible grasp subject to a desired affordance. Given visual perception of the environment, the desired affordance, and the pre-trained models for label extraction (see Section~\ref{sec:method_labels}), region extraction (see Section~\ref{sec:method_regions}), and semantic relations (see Section~\ref{sec:method_semantic_relations}) (line~\ref{alg_line:input1} to \ref{alg_line:model2}), the end-to-end execution is as follows. First, the visual data is processed to extract the labels and map them into a binary vector, where the non-zero entries indicate the presence of an attribute (line~\ref{alg_line:dcnn}) and the feature-label map (line~\ref{alg_line:svm}) describing the object to manipulate. The extracted labels are used to define an \ac{FOL} world and query the \ac{KB} model, thus inferring a set of grasp affordance relations \textit{GA\_r} (line~\ref{alg_line:kb}). \textit{GA\_r} indicates the highest affordance probability per grasping region. From this set and given the desired \textit{affordance}, the framework calculates the most suitable \textit{region\_label$^*$}. If the \textit{affordance} is not chosen, the framework selects the affordance corresponding to the highest probability in the set \textit{GA\_r} (line~\ref{alg_line:optimal_label}). The optimal \textit{region\_label$^*$} is then projected to the object \ac{3-D} data using the extracted features-labels map $\textit{3D\_region}$ (line~\ref{alg_line:optimal_region}). On this optimal grasping patch \textit{G$^*$} we calculate a grasping configuration \textit{$^W$\!\!ee\_pose$^*$} in world coordinates (line~\ref{alg_line:ee_pose}) to be sent to the robot (line~\ref{alg_line:send_ee_pose}).

        \begin{algorithm}[t!]
            \caption{end-to-end execution \label{alg:framework_kb}}
            \textbf{Input:} \\
            $\text{CP}$: camera perception. \\ \label{alg_line:input1}
            \textit{affordance}: affordance choice. \\
            $\texttt{imageToLabel}$: DCNN learned model. \\ \label{alg_line:model1}
            $\texttt{regionsFromCloud}$: SVM learned model. \\ \label{alg_line:model2}
            $\texttt{semanticRelation}$: KB learned model. \\

            \Begin{
                \textit{2D\_labels} $\gets$ \texttt{imageToLabels}(CP.\textit{2D\_image}) \\
                \label{alg_line:dcnn}
                
                $\textit{3D\_region} \gets$ \texttt{regionsFromCloud}(CP.\textit{3D\_image})\\ \label{alg_line:svm}
                
                \textit{GA\_r} $\gets$ \texttt{semanticRelation}(\textit{2D\_labels})\\
                \label{alg_line:kb}
                
                 \textit{region\_label$^*$} $\gets$ \texttt{selectGrasp}(\textit{GA\_r},$\;$\textit{affordance})\\
                \label{alg_line:optimal_label}
              
                \textit{G$^*$} $\gets \textit{3D\_region}$(\textit{region\_label$^*$})\\ \label{alg_line:optimal_region}
                
                \textit{$^W$\!\!ee\_pose$^*$} $\gets$ \texttt{extractGraspPose}(\textit{G$^*$})\\ \label{alg_line:ee_pose}
                
                \texttt{sendToRobot}(\textit{$^W$\!\!ee\_pose$^*$}) \label{alg_line:send_ee_pose}
            }
        \end{algorithm}
\section{Evaluation on Grasp Affordance Dataset}\label{sec:experiments}

We evaluate our methodology on a PR2 robotic platform, in both simulated and real-world scenarios. The \ac{2-D} data is perceived with the robot's left \ac{2-D} camera and the \ac{3-D} information with a kinect mounted on its head. We use the end-to-end execution framework as presented in Algorithm~\ref{alg:framework_kb}.

\subsection{Data Collection\label{sc:gathering}}
This work pursues a multi-target prediction of grasp affordances on an object for which the training data needs to be diverse, accurate and consistent. However, when learning object affordances for robotic grasps, one of the greatest obstacles is the lack of datasets that offer a multi-grasp affordance relation.
Therefore, we build a new dataset with highly correlated information that encourages the creation of more robust robotic grasping methods. We use this collected dataset to ensure the reliability of the obtained grasp affordance regions from the \ac{KB}.
We choose $30$ different objects that are commonly found in online datasets used by grasping \cite{bohg2010learning,Lenz2015}, object recognition \cite{wohlkinger20123dnet} and object-location association methodologies \cite{quattoni2009recognizing,kolve2017ai2}, resulting in a prior for grasping, object category, and possible location labels.
Specifically, the pre-defined grasping regions are taken from \cite{jiang2011efficient}.
Nonetheless, these datasets do not relate to each other, and the affordance relation is still unsolved. Consequently, we design a detailed questionnaire containing these different $30$ objects with their corresponding label priors alongside descriptions of visual and categorical attributes, as suggested in \cite{farhadi2009describing}, as well as possible affordances and indoor locations. This questionnaire was presented to a total number of $1{,}269$ subjects. The collected data led to the creation of a total of $3{,}280$ possible worlds that were used in training and testing.
These possible worlds were composed of three visual attributes, each with at least four possible values, eight possible object categories (i.e., the $30$ objects organised as food, electronics and others), seven possible indoor locations (such as kitchen, office and others) and fourteen possible affordable actions closely related to at least three possible grasping regions.
\begin{figure}[t!]
  \centering\includegraphics[width= 8.7cm]{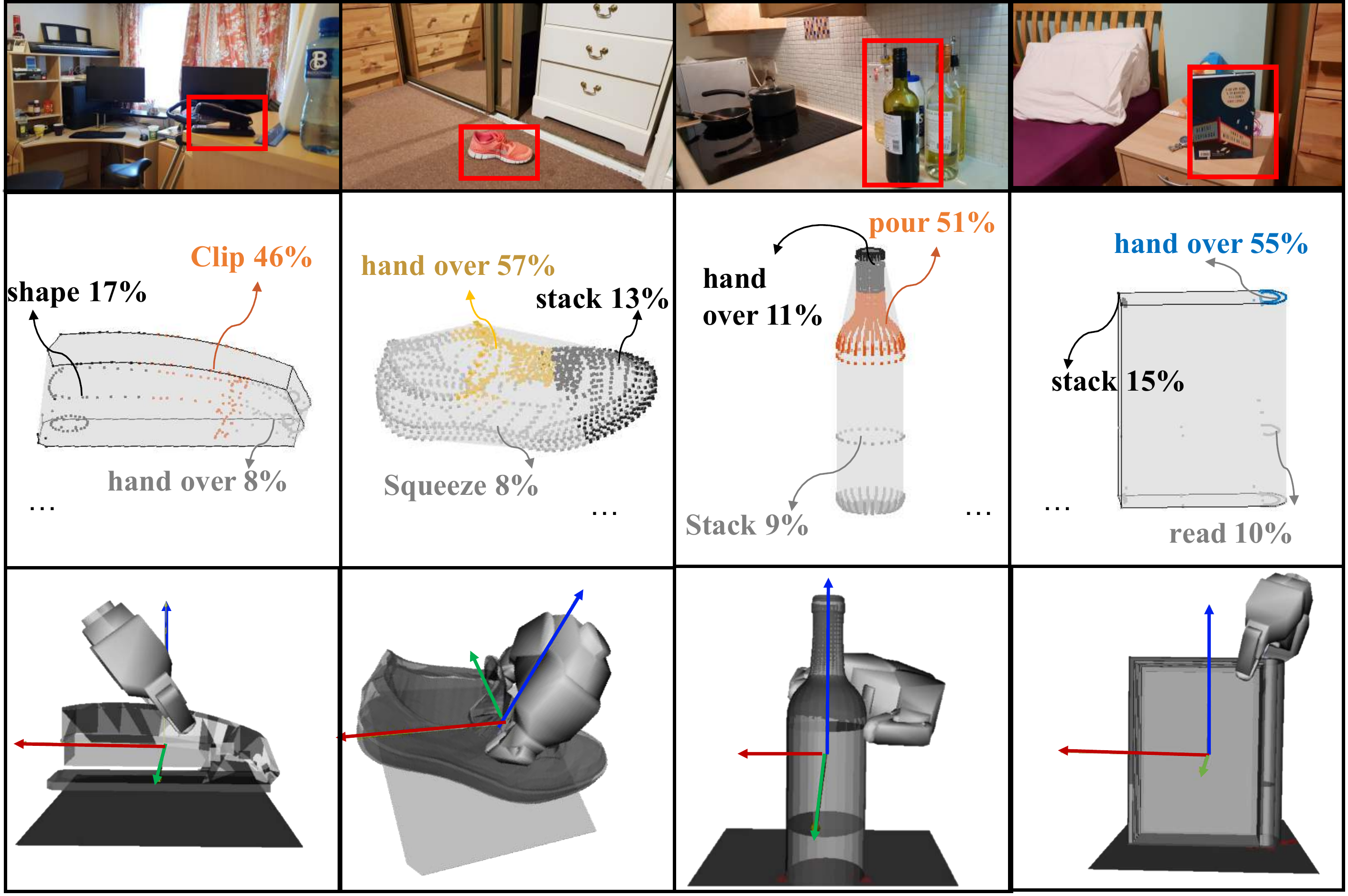}
  \caption{The first row shows an example of an image containing an object and location. The second row is the segmented and processed object \ac{3-D} point cloud with the three highest grasp affordance predictions. The third row is one of the proposed grasp configurations on the obtained grasp affordance region.
  \label{fig:examples_grasps}}
  \vspace{-0.7cm}
\end{figure}
\subsection{Baselines for Grasp Affordances Evaluation\label{ssc:baselines}}
As explained in Algorithm~\ref{alg:framework_kb}, our method is able to evaluate different sets of grasps and select the one that will maximise the success of an affordance. For the following set of experiments, we query the \ac{KB} not only for the grasp region but also for the affordance prediction, as the query shown in Table~\ref{tb:examples_query}, to perform a complete evaluation.
First, we extract the attributes that build the \ac{KB} queries for inference. We use $30\%$ of the objects for testing. These objects are semantically similar to the $70\%$ of objects used for training. To predict their affordances, we take ten images per object in different environments. Given a \ac{2-D} image we extract the scene location, visual and categorical attributes describing the object using a deep \ac{CNN}. Second, we collect the scores from the binary vector (i.e. non-zero entries).
Finally, these attributes are translated into relational worlds using \ac{FOL} and input in the \ac{KB} where we query the grasp affordance regions.
Fig.~\ref{fig:examples_grasps} shows examples of the images taken for the grasp affordance prediction. The first row contains the \ac{2-D} image of the environment with the object. The second row depicts the relevant \ac{3-D} data on the extracted shape context with the top three highest grasp affordance probabilities, and the third row is the resulting grasp configuration on $G^*$.

\subsection{Metrics for Grasp Affordances Regions\label{ssc:metrics}}
\begin{figure}[t!]
  \centering
  \includegraphics[width=6cm]{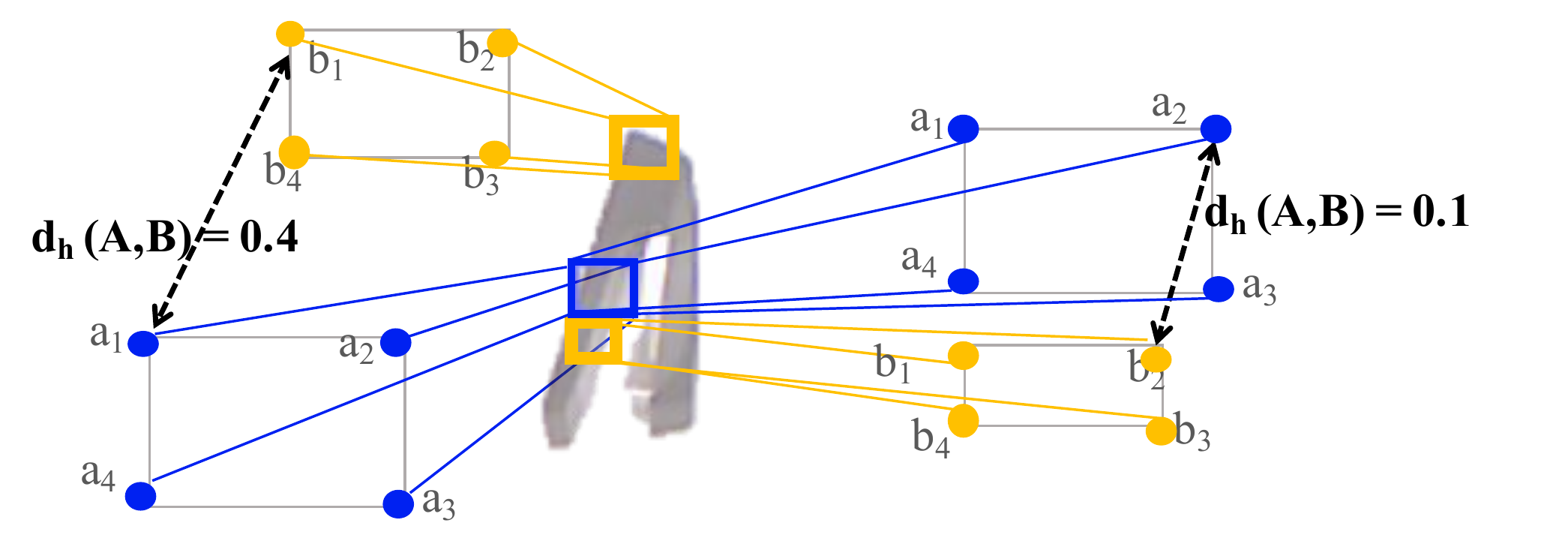}
  \caption{Hausdorff distance example between two grasp regions on a stapler image from \cite{jiang2011efficient}. The grasp affordance rectangle is shown in blue while the ground truth labels are in yellow. The less similar the sets the higher the $d_h$ value. \label{fig:hd_diagram}}
  \vspace{-0.5cm}
\end{figure}

To establish the reliability of our grasp affordance regions, we evaluate the obtained patches against the adopted grasping labels from \cite{jiang2011efficient}. In \cite{jiang2011efficient}, the label is a rectangle that covers the grasping area with the corresponding \ac{2-D} and \ac{3-D} mappings. We choose four instances per object belonging to the $30\%$ of the testing data from \cite{jiang2011efficient} and simulated to be in an office environment. On the \ac{2-D} data, we project the ground truth and our grasp affordance regions are enclosed in rectangles.
Methods that have used the same database to extract the grasping labels \cite{bohg2010learning,Lenz2015} have based their evaluation on measuring the Euclidean distance between rectangle centroids.
As we generalise a segment of an object as a grasping patch, these metrics might overestimate the performance of the algorithm (if one ground truth rectangle is inside a large obtained grasp affordance rectangle), or underestimate it (if the obtained grasp affordance region does not intersect but is close). Thus, we use the Hausdorff distance as a metric of choice to establish the similarity between the two projected rectangles set. The Hausdorff distance is the maximum of all distances from a point in one set $A$ to the closest point in another set $B$, the smaller the value, the more similar the sets are, i.e.:
\begin{equation}
    d_h(A,B)=\max_{a \in A}(\min_{b \in B} d(a,b)),
\end{equation}
where $a$ and $b$ are points in sets $A$ and $B$ respectively, and $d(a,b)$ is the Euclidean distance between $a$ and $b$. Fig.~\ref{fig:hd_diagram} shows how the Hausdorff distance accurately measures the similarity between rectangle patches, although they do not intersect.

\section{Experiments and Discussion}\label{sec:results}

\begin{figure*}[th!]
    \vspace{-0.35cm}
        \centering
        \subfloat[Cubic-like objects\label{fig:cubic}]{
            \centering    \includegraphics[width=7.5cm]{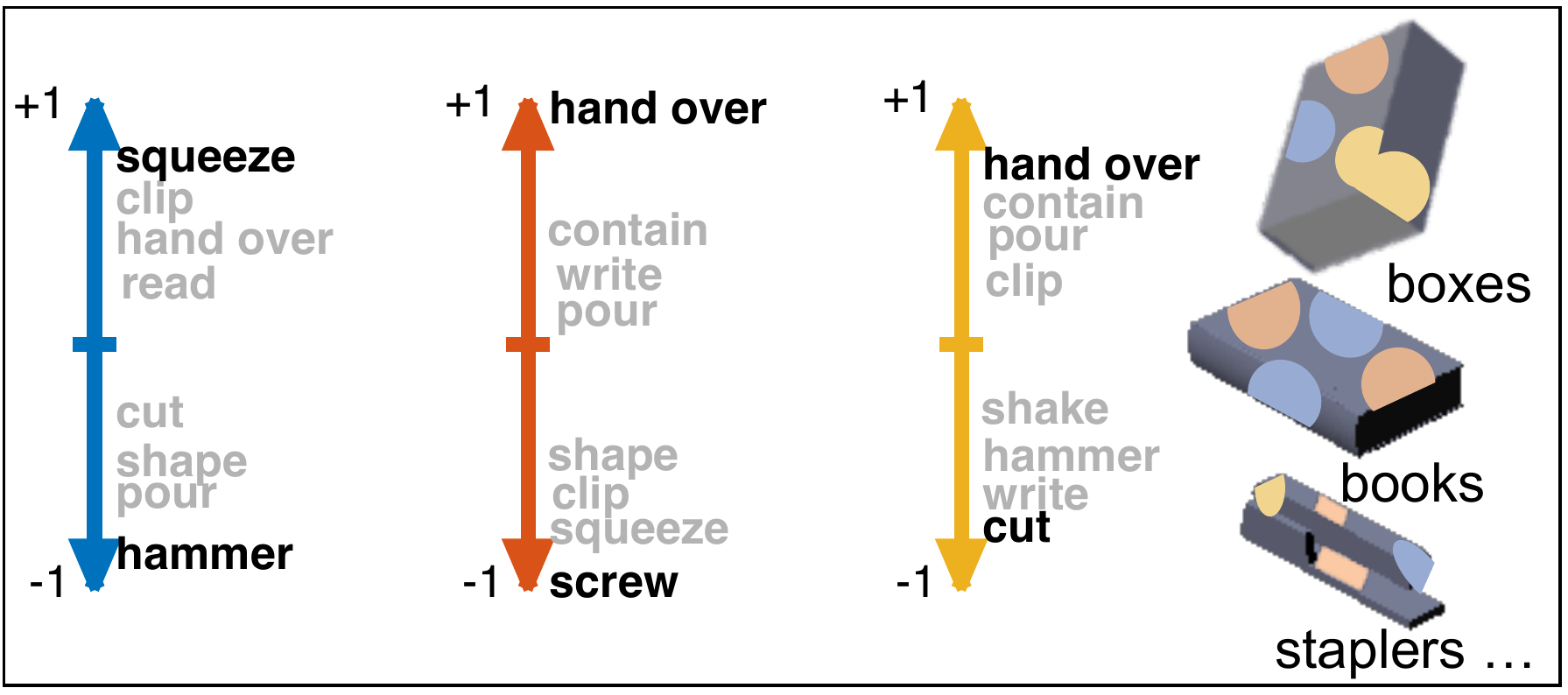}
        }
        \subfloat[Cylindrical-like objects\label{fig:cylindrical} ]{
            \centering
                \includegraphics[width=7.5cm]{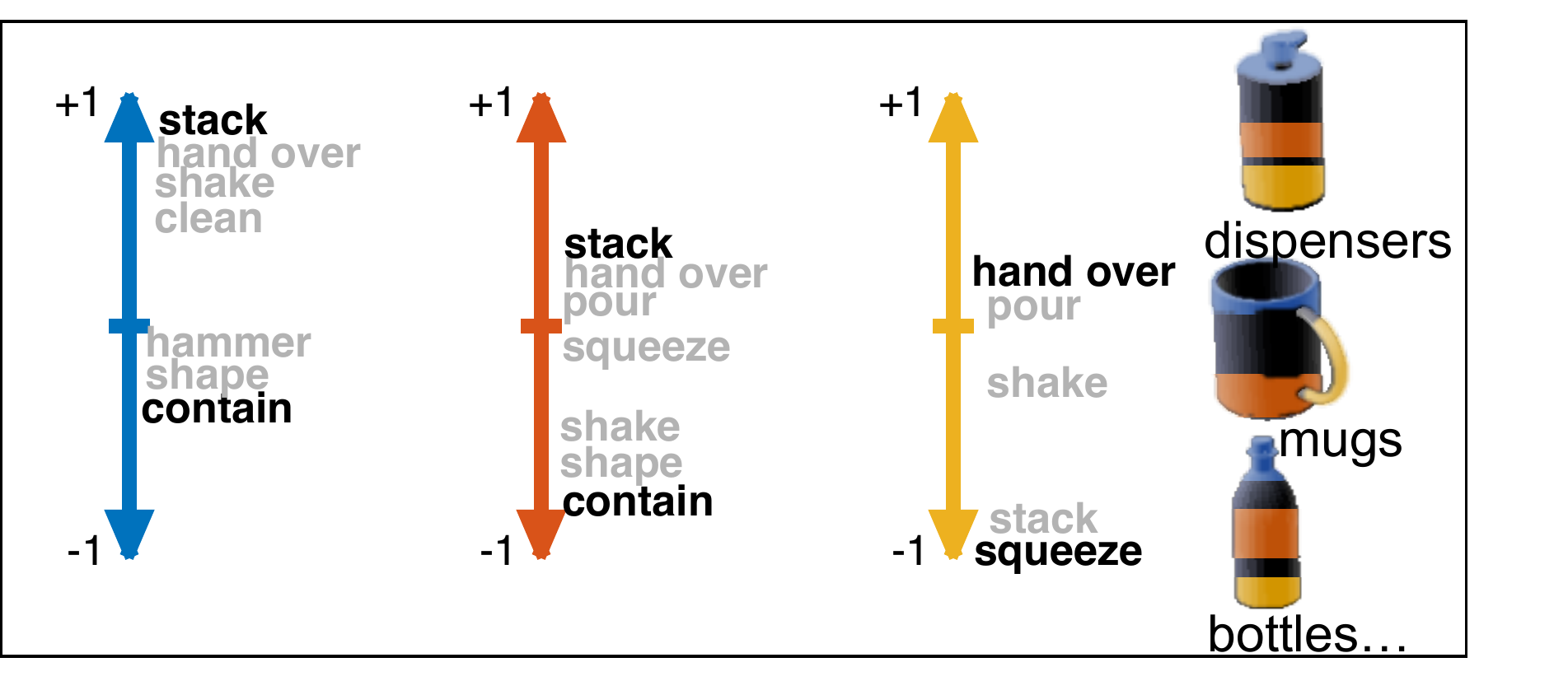}
        }
      \vspace{-0.3cm}
        \subfloat[Irregular-like objects\label{fig:elongated}]{
            \centering
        \includegraphics[width=7.5cm]{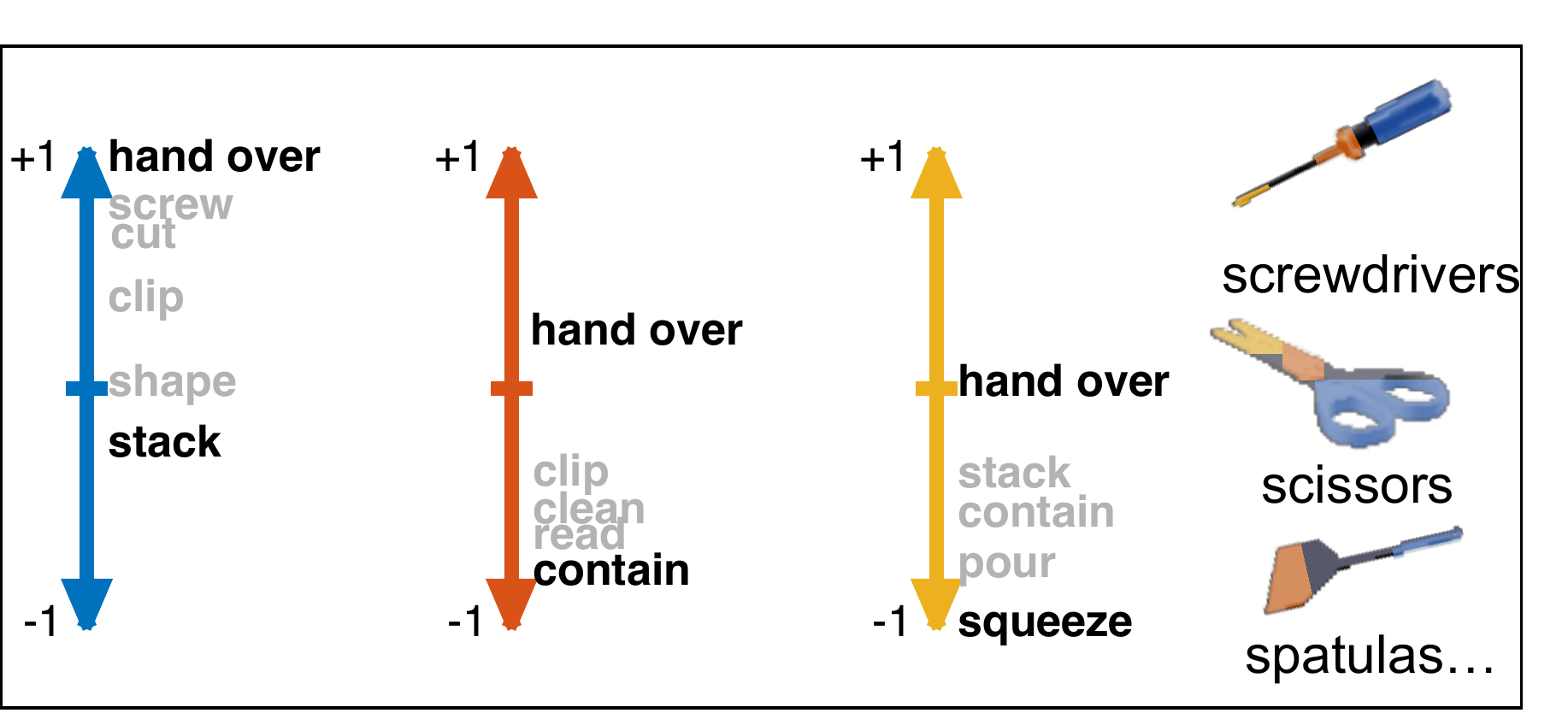}
        }
        \subfloat[Spherical-like objects\label{fig:circular}]{
            \centering
        \includegraphics[width=7.5cm]{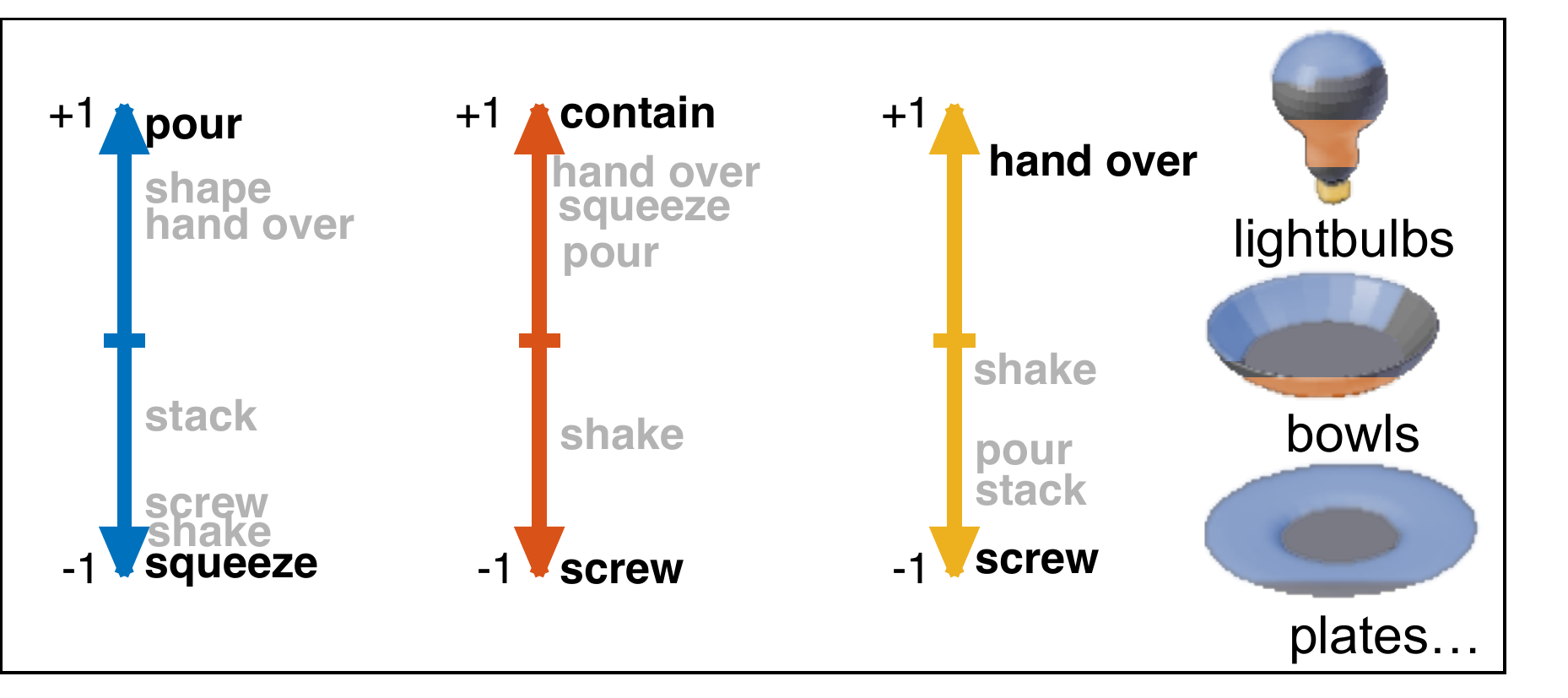}
        }

            \caption{Visualisation of the normalised grasp affordance likelihood learned in Section~\ref{sec:method_semantic_relations} subject to objects' shape context ((a)-(d)) and grasp regions (colour coded arrows with corresponding regions on the objects). The more positive the weight, the more likely that region offers a feasible grasp for the indicated affordance. We include the likelihoods close to the extremes.}
            \label{fig:shape_context}
    \vspace{-0.5cm}
    \end{figure*}

The goal of this work is to reason about the feasible grasps for an object given an affordance. Thus it is important to (i)~test the accuracy of grasp affordance prediction, and (ii)~test the reliability of the grasps in changing scenarios\footnote{More experiments can be found in \url{https://youtu.be/aaA3NA-S5KY}.}.

\subsection{Zero-Shot Grasp Affordance\label{sc:zero}}
Our first evaluation tests the performance of the \ac{KB} on unseen scenarios. In contrast to \cite{zhu2014reasoning}, our feature extraction approach is a deep \ac{CNN} architecture that extracts the objects attributes. We use eight objects semantically similar to the ones used in training, as explained in Section~\ref{ssc:baselines}. We evaluate the performance of our proposed \ac{KB} against three state-of-the-art reasoning methodologies: (i)~a \ac{KB} built with a series of L1-regularised logistic classifiers \cite{farhadi2009describing}, (ii)~our \ac{KB} based on decision trees \cite{ardon2018Towards} and, (iii)~the \ac{KB} based on \ac{MLN} proposed in \cite{zhu2014reasoning} with their \ac{SVM} ranking function for feature extraction.
Table \ref{tb:KB_terminology} shows the mean \ac{AUC} under the \ac{ROC} curve over all the possible grasp affordances.
The results show that our method has the best performance of all methods tested since we train our \ac{KB} over a combination of highly correlated object predicates and relevant constant terms.
Additionally, we note the improved performance of \acp{KB} that use \ac{MLN} over the ones trained with a battery of classifiers.
By using \ac{MLN}, the attributes build relationships regarding object grasp affordances that the classifiers fail to incorporate.

\subsection{Grasp Affordances Relation\label{sc:relation}}

\begin{figure}[th!]
  \centering
  \includegraphics[width= 8.5cm]{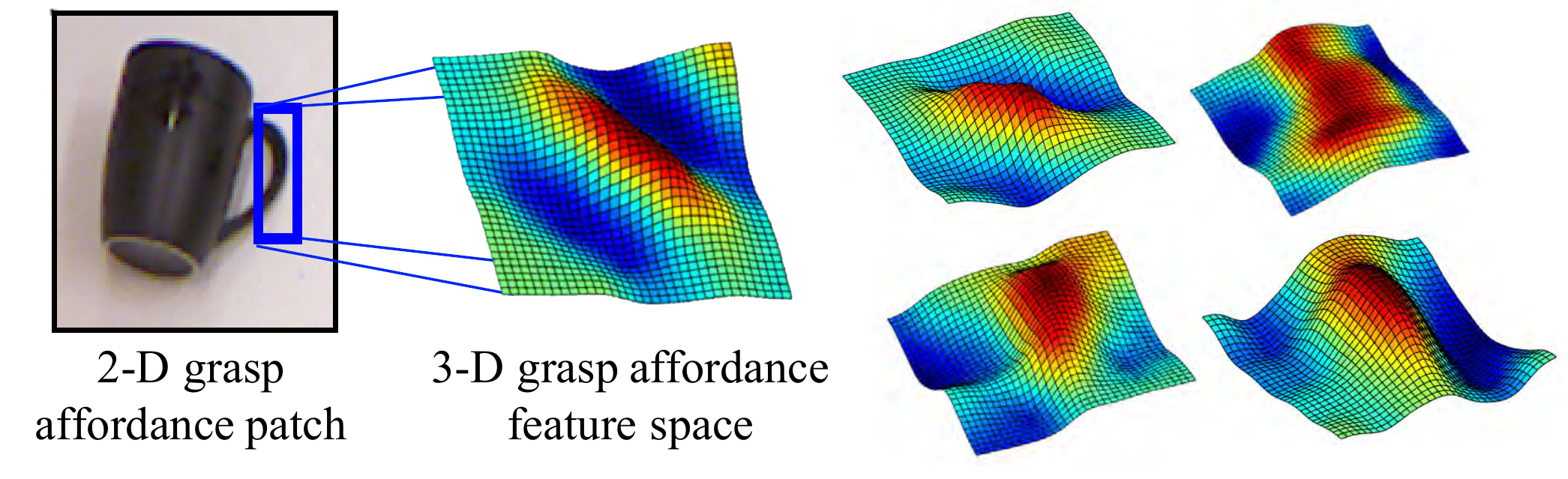}
  \caption{Examples of learned grasp affordance patch features of affordances such as \textit{hand over}, \textit{stack} and \textit{pour}.\label{fig:features}}
\end{figure}

Our primary contribution is to associate a set of grasp affordances with an object. Fig.~\ref{fig:shape_context} portrays possible grasp affordances for objects in different shape contexts across different indoor scenes. The three arrows represent the three grasping regions across different objects with more affordance possibilities. The regions are colour coded on the corresponding area of the different objects. The affordances are sorted by the normalised weights between -1 to +1 per grasping region, where the higher the weight, the more likely that affordance is to be successful when grasping the object using the colour coded grasping region.
We group the objects by shape context for a clearer grasp affordance representation. Out of the $14$ possible affordances, we extracted those with higher and lower relational weight. Among the different grasp affordances, one of the most probable ones across shape contexts is object \textit{hand over}. Specifically, this is the case for objects that are used as tools and are recognised with an irregular shape (Fig.~\ref{fig:elongated}). Because the \ac{KB} has learned from data collected from humans, it reflects the likelihood of that grasp affordance region to be more or less ``acceptable" than others for a particular action. Also in Fig.~\ref{fig:elongated}, the likelihood of success at \textit{handing over} the objects from the grasp region indicated in blue (i.e., object handles) is higher than the other two. The same case is shown in Fig.~\ref{fig:cubic} and Fig.~\ref{fig:cylindrical} for \textit{stack}. Moreover, Fig.~\ref{fig:features} illustrates some of the grasp affordance feature patches of the \textit{hand over}, \textit{stack} and \textit{pour} affordances learned with our method. These patches (specifically red areas) correspond to graspable regions of objects such as handles or other raised regions.

\begin{table}[b!]
    \renewcommand{\arraystretch}{1.25}
    \centering
     \begin{tabular}{l|c|c|c}
     \hline
     \multirow{2}{*}{\textbf{Approach}} & \multicolumn{3}{c}{\textbf{Performances per attribute ($\overline{\text{AUC}}$)}} \\ \cline{2-4}
  &\textbf{visual} &  \textbf{categorical} &  \textbf{all+location}  \\ \hline
  L1-LR\cite{farhadi2009describing} & 0.70& 0.74 & 0.77\\ \hline
  \ac{SVM} \ac{KB} \cite{zhu2014reasoning} & 0.73 & 0.77 &0.82\\ \hline
  our previous \ac{KB} \cite{ardon2018Towards} & 0.72 & 0.75  & 0.79  \\ \hline
 \textbf{our \ac{KB}} &  0.75 & 0.79  & \textbf{0.84} \\ \hline

     \end{tabular}
     \caption{Performance of Zero-Shot grasp-affordance prediction for different attributes and their combination.\label{tb:KB_terminology}}
 \end{table}

\subsection{Grasp Affordance Selection Reliability\label{sc:reliability}}

We demonstrate the reliability of our grasp affordance hypothesis by (i)~evaluating the accuracy of the affordance detection, and (ii)~checking if the obtained grasping regions correspond to stable grasps using Hausdorff distance. We train two state-of-the-art methods, using deep learning \ac{CNN} methodologies, with our dataset and evaluate separately (i) and (ii) given that the current literature does not treat the grasp affordance as a unified task.
\subsubsection{Affordance Detection} We compare our affordance detection with \cite{AffordanceNet18} (\textbf{Detection (\%)} in Table~\ref{tb:grasp}).  We use a total of $64$ images of $16$ object classes in different environments, among which the eight used for zero-shot prediction are included.
Our method shows a higher performance, by using \ac{MLN}, since we build a series of relationships around an object (\textit{worlds}) that traditional machine learning methods fail to connect. Namely, \cite{AffordanceNet18} does not consider the task given a context. Instead, it learns object part labels and categories to assign an affordance. Fig.~\ref{fig:affordance_context} shows an example of a knife grasp affordance detection. The knife affords equally two tasks when using \cite{AffordanceNet18}, while the affordances are correct (\textit{cut and hand over}), when grasping, determining one task is essential.
\subsubsection{Similarity between patches}
\begin{figure}[t!]
\vspace{-0.7cm}
  \centering
      \includegraphics[width=7.5cm,trim = 0cm -1cm 0cm 0cm]{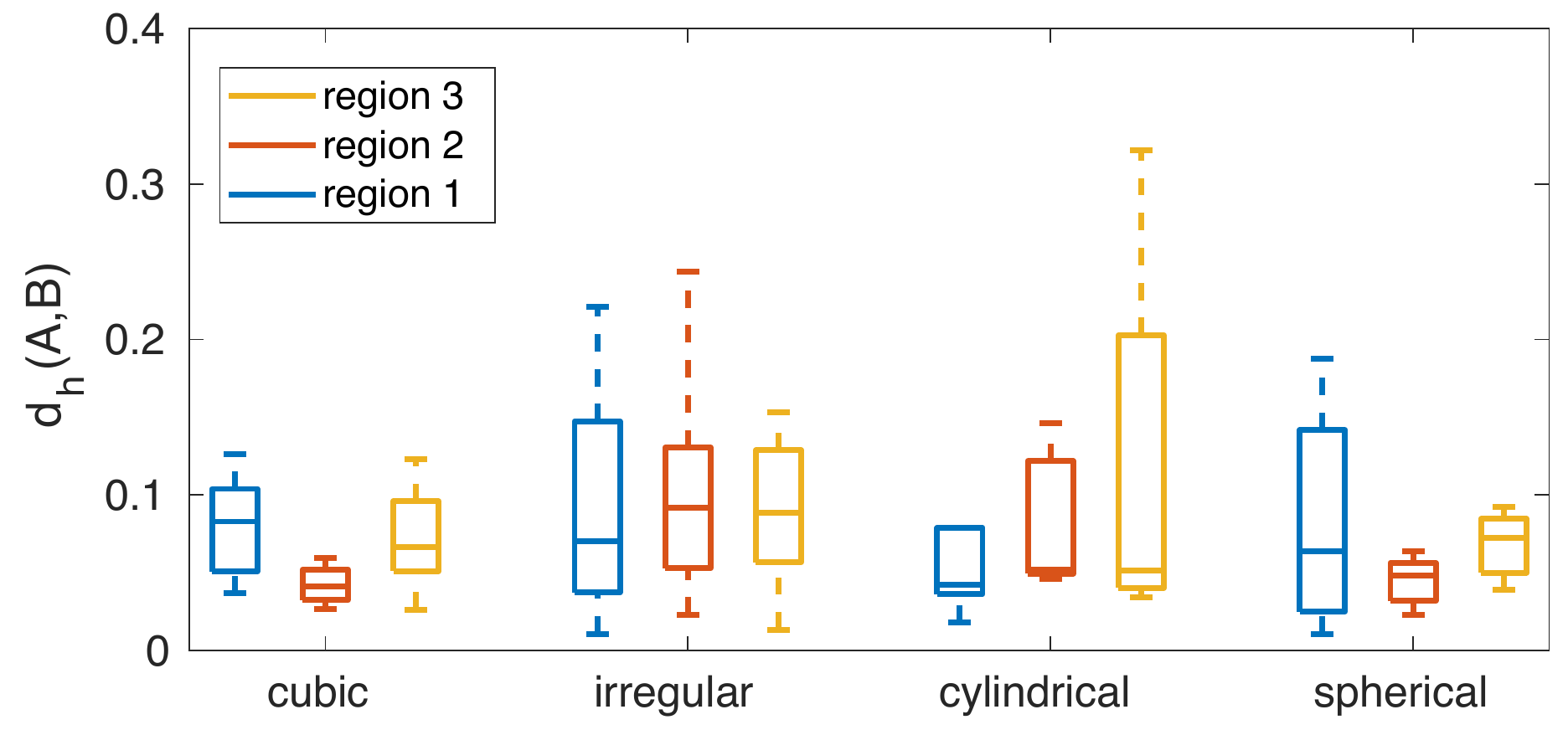}
      \vspace{-0.6cm}
  \caption{Hausdorff distance to measure the similarity between ground truth regions and our grasp affordances.\label{fig:object_samples_for_zero_shot}}
\end{figure}

First, we check the similarity between our regions and the original ones from~\cite{jiang2011efficient}. We use the subset of $32$ images (Section~\ref{ssc:metrics}) to project the ground truth and our obtained areas. Fig.~\ref{fig:object_samples_for_zero_shot} shows the Hausdorff distance, $d_h$ between the obtained hypotheses, set $A$, and the ground truth labels, set $B$, grouped by shape context and grasping regions. The $d_h$ mean per grasp affordance region are below $0.1$ for all the objects. Specifically, a low $d_h$ is obtained for rectangles that are nearby ($d_h \le 0.1$) while larger values might be obtained by far apart sets ($d_h \ge 0.4$) (Section~\ref{ssc:metrics}).
Second, we compare our method with \cite{Lenz2015} which finds multiple reliable grasping regions on the data in \cite{jiang2011efficient}. We use the subset of $64$ images and compare the Hausdorff distance. Table~\ref{tb:grasp} shows that both methods achieve considerably small and similar $d_h$, thus learning stable grasping patches. Examples of grasping patches obtained with both methods, ours and \cite{Lenz2015}, are shown in Fig.~\ref{fig:affordance_context}.

\subsection{Grasp Affordance on a Robotic Platform\label{sc:robotic}}

\begin{figure*}[ht]
        \centering
        \vspace{-0.3cm}
        \subfloat[Method comparison \label{fig:affordance_context}]{
            \centering
                \includegraphics[height=5.1cm]{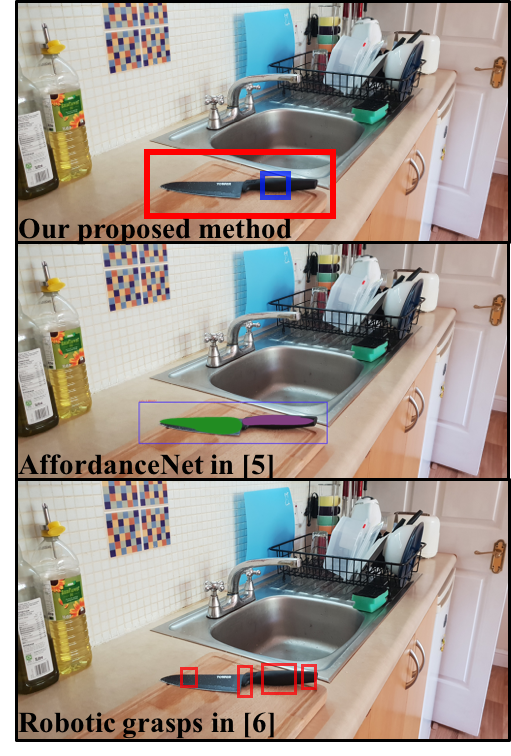}
        } \hspace{-0.25cm}
        \subfloat[Kitchen \label{fig:kitchen}]{
            \centering
        \includegraphics[width=4.55cm]{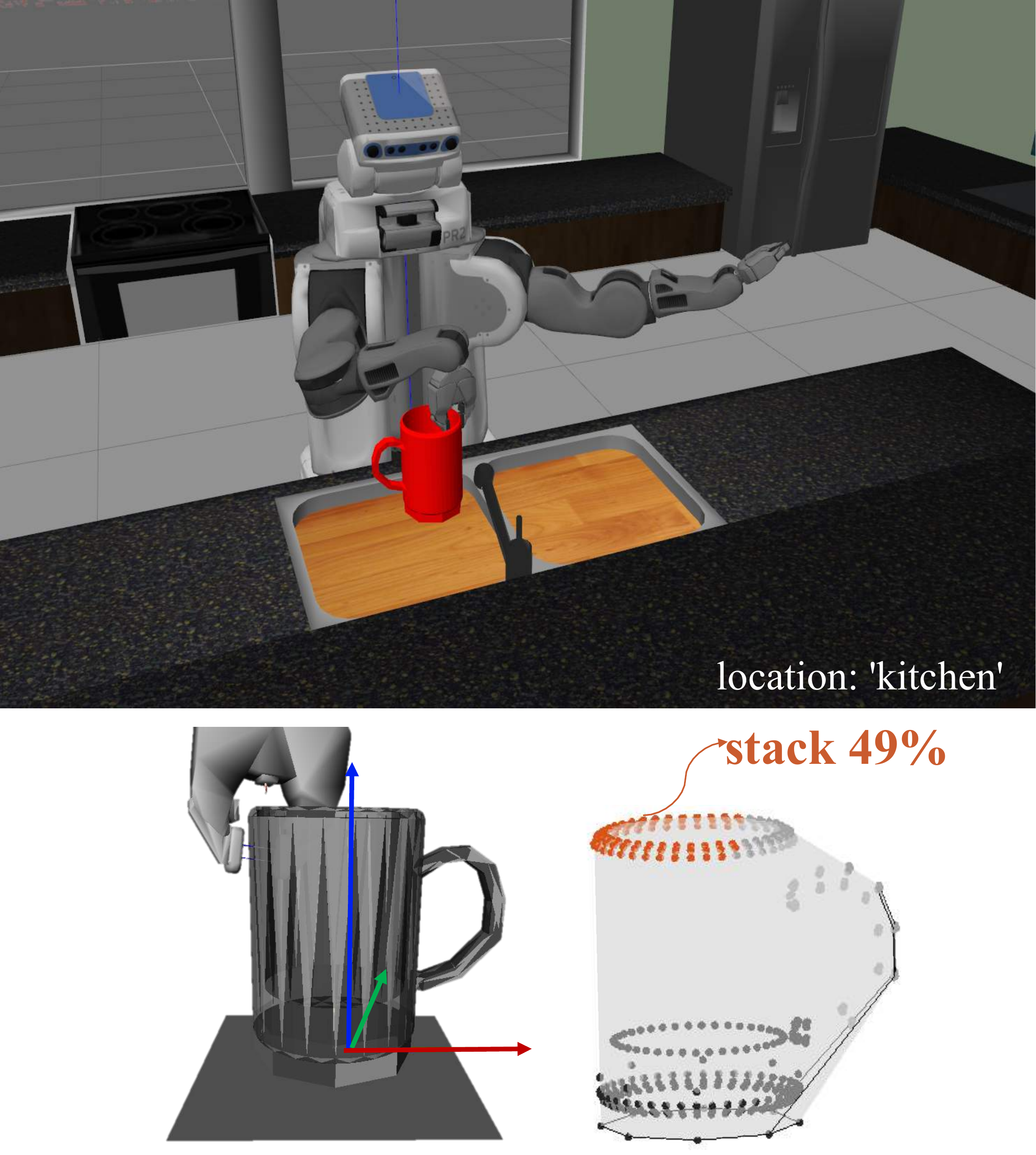}
        } \hspace{-0.25cm}
        \subfloat[Dinning room \label{fig:dinning_room}]{
            \centering
        \includegraphics[width=4.55cm]{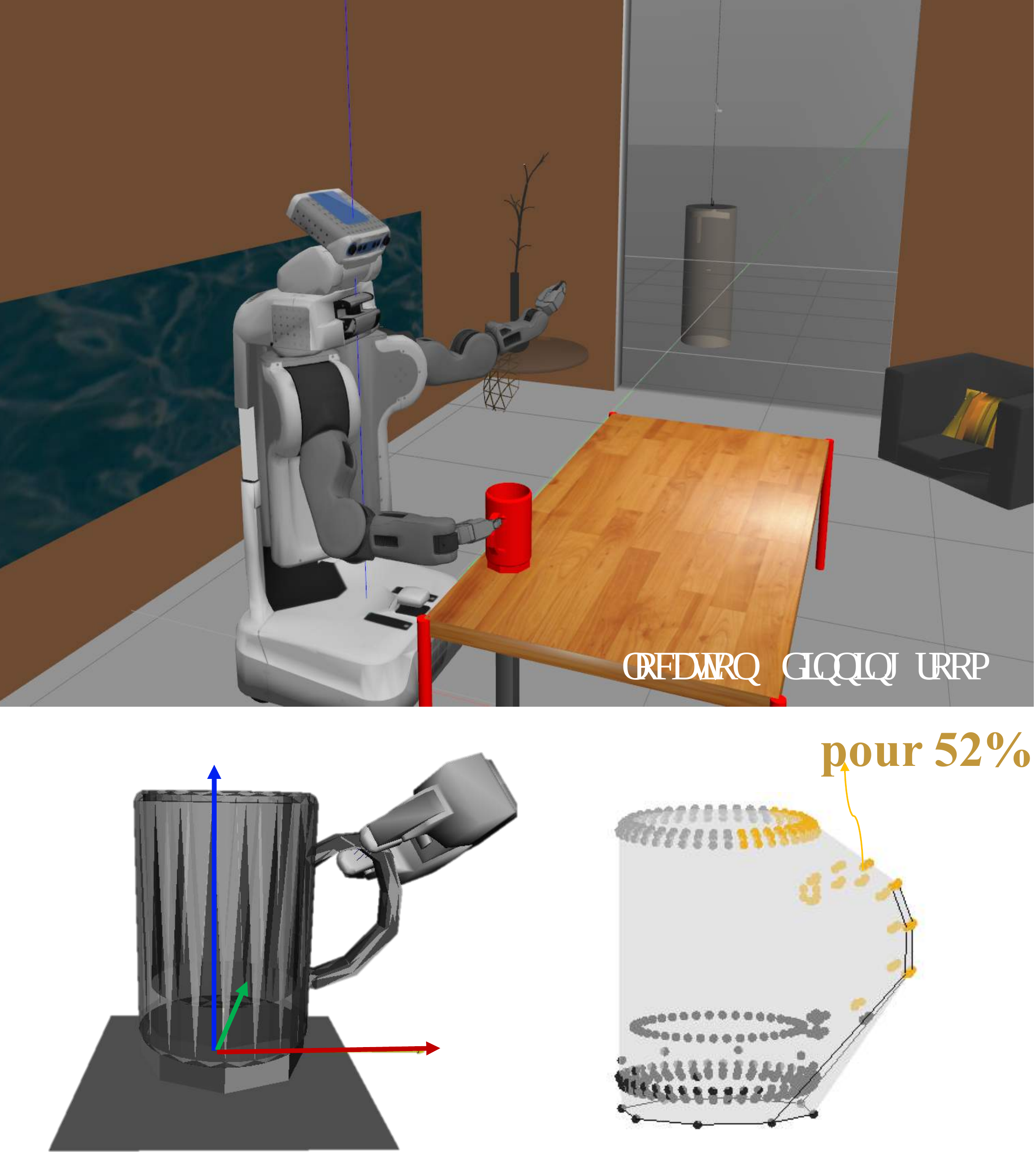}
        }
        \subfloat[Office \label{fig:office}]{
            \centering
        \includegraphics[width=4.55cm]{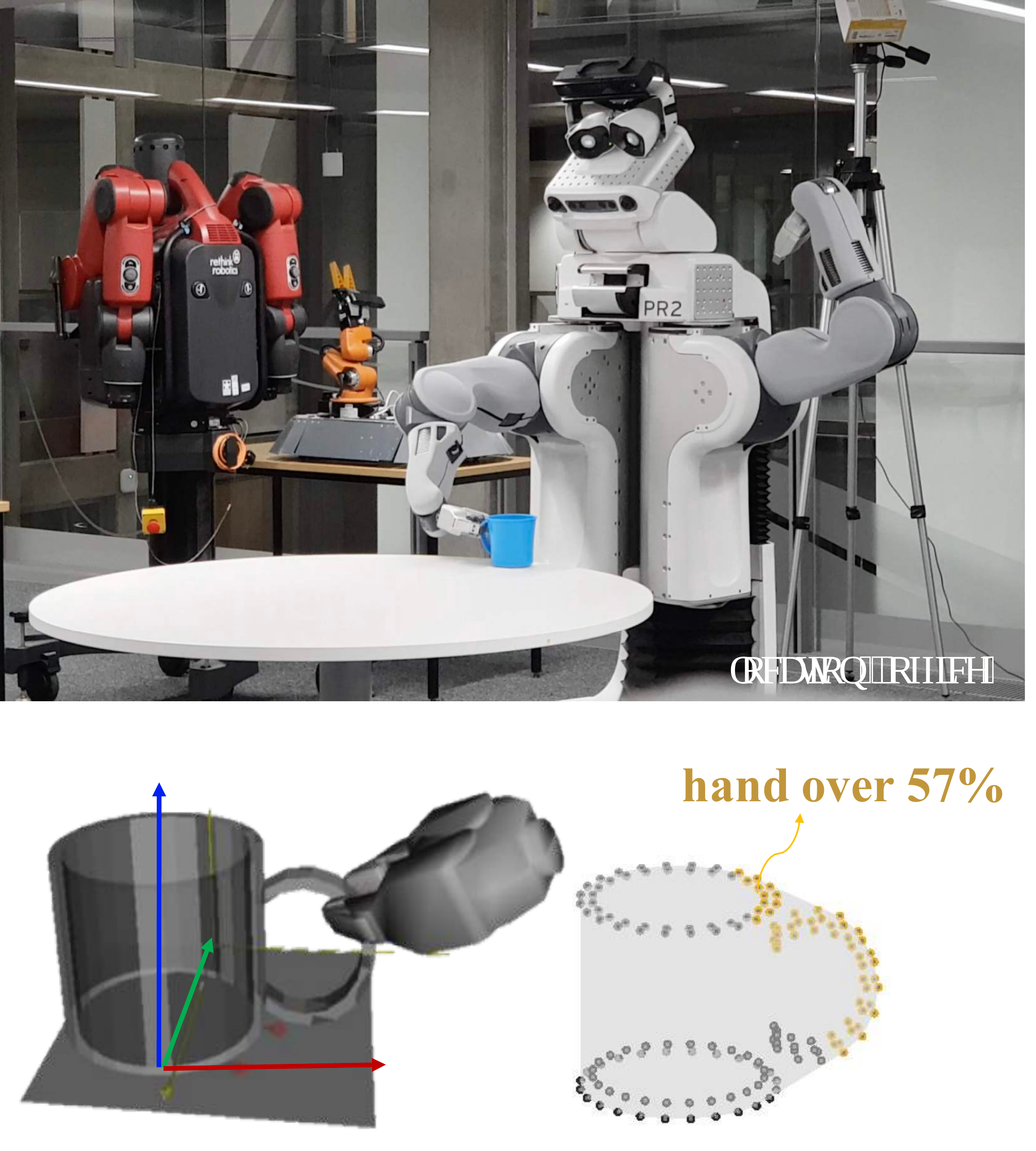}
        } \hspace{-0.25cm}
            \caption{(a)~Comparison of our method with state-of-the-art alternatives~\cite{Lenz2015,AffordanceNet18}, and (b)-(d)~our method running in PR2 to grasp an object in different simulated and real-world scenarios while checking the variations on the detected grasp affordances.}
            \label{fig:examples_grasp}
           \vspace{-0.5cm}
    \end{figure*}

To explore the generalisability and effect of the robot on the success rate of our method, we ran an extensive number of experiments on both a simulated and real PR2 robotic platform. We tested our method in three different indoor scenarios: a kitchen and dining room setting in simulation and a real office environment. For this experiment, we use the previously selected $16$ different object classes, among which we assess robustness by variating instances, and try the affordance detection and grasping task $25$ times per object class for a total of $400$ evaluations. The performance of the grasp affordance task is detailed in Table~\ref{tb:grasp} (\textbf{On the robot (\%)}).
We average the grasp affordance prediction with the actual grasping action success on the correct detection cases (\textbf{avg} in Table~\ref{tb:grasp}). The grasp is considered successful if: (i)~the gripper approaches the grasp affordance region of the object below a Hausdorff distance threshold ($d_h < 0.2$), and (ii)~if the object is successfully grasped. The lowest performance is obtained by objects with irregular shape context where the most significant setback is the success of the grasping action given that, in general, the set of objects were too small for the PR2 gripper; thus they frequently slip.
Fig.~\ref{fig:kitchen}-\ref{fig:office} illustrate the experimental set-up for the three different scenarios, focusing on the grasp affordances of a mug. Interestingly, the affordances detected on the object in the three locations are different. In two out of the three scenarios, dining room and office, the grasping region coincides even though the affordances are different. The method relates the location with object semantics to decide on a grasp affordance that will potentially ensure the accomplishment of an action. Given that the \ac{KB} learned from our collected data, it reflects what is socially ``acceptable" in an environment. Although the grasping region for the mug is the same in the dining room and the office, in the office it is more likely \textit{hand over} the mug than to \textit{pour} liquids from it.

\begin{table}[b!]
    \renewcommand{\arraystretch}{1.25}
     \centering
     \begin{tabular}{p{1.1cm}|c|c|c|c|c|c}
     \hline
     \multirow{2}{0.6cm}{\centering\textbf{Objects shape}} &
     \multicolumn{2}{c|}{\textbf{Detection (\%)}}  &
     \multicolumn{2}{c|}{$\boldsymbol{d_{h}}$} &   \multicolumn{2}{c}{\centering{\textbf{On the robot (\%)}}}  \\ \cline{2-7}
  &  \textbf{\cite{AffordanceNet18}} & \textbf{ours} & \textbf{\cite{Lenz2015}} & \textbf{ours} & \textbf{detection/grasp} & \textbf{avg}\\ \hline

  Cubic  & 82.7 & 88.5 &0.05 &0.01& 90.3 / 100 & 95.2\\ \hline
  Cylindrical  & 79.6  & 87.4 &0.01 &0.03 & 87.1 / 96.1 & 91.6 \\ \hline
  Irregular  & 77.8 & 88.6 & 0.09 & 0.12  & 87.9 / 77.3 & 82.6 \\ \hline
  Spherical & 83.2 & 90.5 & 0.02 & 0.03  & 91.6 / 100 & 95.8\\ \hline

     \end{tabular}
     \caption{Comparison with state-of-the-art methods and grasp affordance performance of the robotic platform.\label{tb:grasp}}
 \end{table}

\section{Conclusions and Future Work}\label{sec:contribution}

We presented a new method for reasoning about the different grasp affordances of an object.
In contrast to state-of-the-art techniques, instead of hand-defining the grasp affordance labels on the objects, we collected data from $1{,}269$ different participants to obtain their input on the relation of object attributes, locations and grasp affordance labels.
Using this collected data, our approach not only learns grasp affordances but also learns to characterise socially acceptable grasp behaviours on different objects in various scenarios.
The information included in this dataset opens doors in the research community towards more robust and heterogeneous robotic grasping methods.
The proposed method also outperforms alternative grasp affordance recognition techniques. We attribute this performance to our structures for grounding and relating data. Our method is able to (i) reason about the most probable grasp affordance, among a set, by inferring the contextual semantics relation, and (ii) map the optimal grasp affordance to the \ac{3-D} data of the object and proceed with the grasp using a robotic manipulator.
Moreover, this work encourages interesting future studies such as the prediction of action probabilities to be executed by associating objects in the scene, the evaluation of the generalisability of our method with different manipulators, and the assessment of end-state comfort-effect for grasping in human-robot collaboration tasks.

\bibliographystyle{ieeetr}
\bibliography{bibliography}
\end{document}